# PhenoProfiler: Advancing Phenotypic Learning for Image-based Drug Discovery

Bo Li[1], Bob Zhang[1,*], Chengyang Zhang[2], Minghao Zhou[3], Weiliang Huang[1], Shihang Wang[4], Qing Wang[3], Mengran Li[5], Yong Zhang[2], Qianqian Song[3,6,*]

[1]PAMI Research Group, Department of Computer and Information Science, University of Macau, Taipa, Macau SAR, China. [2]Beijing Key Laboratory of Multimedia and Intelligent Software Technology, Beijing Institute of Artificial Intelligence, Beijing University of Technology, Beijing, China. [3]Department of Health Outcomes and Biomedical Informatics, College of Medicine, University of Florida, Florida, USA. [4]Faculty of Applied Sciences, Macao Polytechnic University, Taipa, Macau SAR, China. [5]School of Intelligent Systems Engineering, Sun Yat-sen University, Guangdong, China. [6]Department of Cancer Biology, Wake Forest School of Medicine, Winston Salem, NC, USA.

*Corresponding authors

## ABSTRACT

In the field of image-based drug discovery, capturing the phenotypic response of cells to various drug treatments and perturbations is a crucial step. This process involves transforming high-throughput cellular images into quantitative representations for downstream analysis. However, existing methods require computationally extensive and complex multi-step procedures, which can introduce inefficiencies, limit generalizability, and increase potential errors. To address these challenges, we present PhenoProfiler, an innovative model designed to efficiently and effectively extract morphological representations, enabling the elucidation of phenotypic changes induced by treatments. PhenoProfiler is designed as an end-to-end tool that processes whole-slide multi-channel images directly into low-dimensional quantitative representations, eliminating the extensive computational steps required by existing methods. It also includes a multi-objective learning module to enhance robustness, accuracy, and generalization in morphological representation learning. PhenoProfiler is rigorously evaluated on large-scale publicly available datasets, including over 230,000 whole-slide multi-channel images in end-to-end scenarios and more than 8.42 million single-cell images in non-end-to-end settings. Across these benchmarks, PhenoProfiler consistently outperforms state-of-the-art methods by up to 20%, demonstrating substantial improvements in both accuracy and robustness. Furthermore, PhenoProfiler uses a tailored phenotype correction strategy to emphasize relative phenotypic changes under treatments, facilitating the detection of biologically meaningful signals. UMAP visualizations of treatment profiles demonstrate PhenoProfiler's ability to effectively cluster treatments with similar biological annotations, thereby enhancing interpretability. These findings establish PhenoProfiler as a scalable, generalizable, and robust tool for phenotypic learning, offering transformative advancements for image-based high-throughput drug screening and discovery. PhenoProfiler is available as a Python package at https://github.com/QSong-github/PhenoProfiler, with detailed tutorials demonstrating its application to high-throughput imaging datasets. Additionally, we offer a user-friendly web server at https://phenoprofiler.org, providing seamless access to PhenoProfiler's capabilities without the need for coding.

## INTRODUCTION

In image-based drug discovery[1,2], particularly with techniques like cell painting, learning robust image representations is essential for extracting meaningful insights from complex, high-throughput image datasets. Cell painting involves using multiple fluorescent dyes to label various organelles and cellular components, producing multi-channel images that capture phenotypic changes in response to different drugs and perturbations[3]. These high-dimensional images are rich in information, making automated methods for learning image representations crucial. These representations enable the development of predictive models for drug discovery, allowing for better identification of therapeutic compounds, understanding drug mechanisms, and predicting off-target effects[2–4]. Many laboratories and companies have generated extensive cell painting datasets[5–8], facilitating the identification of phenotypic changes in response to drug treatments and supporting various downstream applications. For example, analyzing changes in cell morphology can provide insights into drug targets and mechanisms of action[9,10], while comparisons of cell morphology before and after treatment can help evaluate drug efficacy and identify promising candidates with significant impacts[8,11]. Additionally, these image representations can be integrated with other data types, such as gene expression profiles[7,12], allowing for comprehensive multimodal analyses that further advance drug discovery and biomedical research[5,13].

The high-dimensional nature of cell painting images often introduces redundancy and noise, necessitating extensive preprocessing steps such as normalization, segmentation, and artifact removal[8,14]. Furthermore, the large-scale nature of these datasets demands substantial computational resources for scalable processing, and the extracted morphological features may lack biological interpretability, making it challenging to directly utilize these images for meaningful analysis. Traditional methods such as ResNet50[15] and ViT[16], both pre-trained on ImageNet[17], offer general solutions for image processing and representation learning. ResNet50 provides hierarchical feature representations through residual connections, while ViT captures long-range dependencies in image data. ImageNet pre-training further enhances their generalizability for computational image analysis across diverse datasets. To address the specific challenges posed by cell painting images, tailored methods such as CellProfiler[18] and DeepProfiler[9] have been developed to extract informative and compact representations of cell morphology. CellProfiler is a versatile, open-source tool that enables biologists to create modular pipelines for high-throughput image analysis, measuring various cellular parameters such as size, shape, intensity, and texture. On the other hand, DeepProfiler leverages deep learning techniques including EfficientNet[19] to process multi-channel images from cell painting assays, resulting in robust single-cell morphology profiles. By transforming complex image data into concise and interpretable representations, these methods enable phenotypic profiling, providing valuable insights into drug effects and cellular perturbations.

Despite these advancements, existing methods for morphological representation learning face several critical limitations, particularly when applied to high-dimensional cell painting images. First, these methods often process multi-channel images by decomposing them into multiple sub-images, resulting in a complex and resource-intensive workflow. This approach typically involves segmenting the images to identify individual cell locations, extracting sub-images for each cell, applying deep learning models to extract features from these sub-images, and finally integrating those features to generate a comprehensive representation of the original multi-channel image. This multi-step process not only increases computational and acquisition costs but also introduces additional sources of error, such as inaccuracies in segmentation and feature integration. Second, these methods rely on drug treatment conditions as classification labels[9,20], which provide limited information for capturing the diversity and complexity of cellular responses. This reliance often results in less biologically meaningful phenotypic representations, as those condition labels may fail to capture subtle morphological changes. Moreover, these labels lack universality and are often specific to certain plates or experimental setups. Consequently, existing models trained on these limited datasets struggle to generalize effectively across diverse experimental conditions, reducing their scalability and applicability. These limitations highlight the need for more streamlined, efficient, and robust method to obtain biologically meaningful representations.

In this paper, we introduce PhenoProfiler, an innovative tool for learning phenotypic representations of cell morphology from high-throughput images. Unlike existing methods, PhenoProfiler functions as an end-to-end framework that directly encodes whole-slide multi-channel images into low-dimensional feature representations, without the need for extensive preprocessing such as segmentation and sub-image extraction. PhenoProfiler consists of three main modules: a gradient encoder, a transformer encoder, and a multi-objective learning module that integrates classification, regression, and contrastive learning. These modules establish a unified and robust feature space for representing cellular morphology. Additionally, PhenoProfiler incorporates a tailored phenotype correction strategy, designed to emphasize relative changes in cell phenotypes under different treatment conditions, enhancing its ability to capture meaningful biological signals. Extensive benchmarking on over 230,000 multi-channel images demonstrates PhenoProfiler's state-of-the-art performance in extracting accurate and interpretable phenotypic representations. By addressing the complexity, high costs, and limited generalization capabilities of existing workflows, PhenoProfiler represents a significant advancement in phenotypic profiling and offers a powerful tool for accelerating image-based drug discovery. PhenoProfiler is available as a Python package at https://github.com/QSong-github/PhenoProfiler. Additionally, we provide a user-friendly web server at https://phenoprofiler.org/, enabling seamless access to its powerful features without requiring coding expertise.

## RESULTS

### Overview of the PhenoProfiler model

PhenoProfiler learns morphological representations and extracts phenotypic changes of treatment effects from high-throughput images. Different with existing methods (**Fig. 1a**), PhenoProfiler is designed as an end-to-end model with three key modules (**Fig. 1b**): a gradient encoder using difference convolution[21,22], which enhances cell edge information and improves the clarity and contrast of cell morphology. A transformer encoder with multi-head self-attention mechanisms[20] further captures long-range dependencies and intricate relationships within the data, ensuring comprehensive feature extraction. A multi-objective learning module, composed of two multi-layer perceptron layers, is then utilized to enhance accuracy and generalization. This module integrates classification, regression, and contrastive learning. The classification learning maps image representations to categorize treatment conditions based on their corresponding labels. The regression learning leverages the rich and continuous supervisory information provided by regression objective to capture detailed morphological representations across different treatments, thereby significantly improving model performance. The contrastive learning improves robustness and generalization by maximizing similarity among representations of similar treatment conditions and minimizing it among dissimilar ones. After well training, PhenoProfiler identifies a unified and robust feature space for representing cell morphology. During the inference phase (**Fig. 1c**), PhenoProfiler utilizes a phenotype correction strategy to emphasize relative phenotypic changes under different treatment conditions, thereby revealing associated biological matches and treatment-associated representations.

### Superior performance of PhenoProfiler in biological matching tasks

To evaluate the performance of PhenoProfiler, we conduct a quantitative assessment in downstream biological matching tasks[9]. Herein, we compare PhenoProfiler with several established models, including DeepProfiler[9], ResNet50[15], and ViT[16] (details provided in the **Benchmarking Methods** section). Two evaluation metrics are used: Folds of Enrichment (FoE) and Mean Average Precision (MAP). FoE measures the model's ability to identify biologically related treatments (e.g., those sharing the same mechanism of action or genetic pathway) among its top predictions, thereby assessing the accuracy of its captured representations[9,23]. MAP evaluates the model's overall ranking performance, balancing precision (the proportion of relevant matches among top predictions) and recall (the proportion of all relevant matches successfully retrieved). This metric captures the model's effectiveness in ranking all biologically related treatments and demonstrates its capability in morphological representation. Detailed descriptions of these metrics are provided in the **Evaluation Metrics** section. For benchmarking, we utilize over 230,000 images from three publicly available cell painting data resources: BBBC022[24], CDRP-BIO-BBBC036[25], and TAORF-BBBC037[26]. These data are collected from 231 plates and include two categories of treatment: compounds and gene overexpression perturbations, encompassing 4,285 different treatments. These diverse and extensive datasets ensure a comprehensive and robust evaluation of the methods.

As illustrated in **Fig. 2a**, PhenoProfiler consistently outperforms other models. For the FoE metric, significant improvements reveal relative gains of 64.9%, 3.4%, and 50.0% for PhenoProfiler over DeepProfiler on the BBBC022, CDRP-BIO-BBBC036, and TAORF-BBBC037 datasets, respectively. The modest improvement observed on the CDRP-BIO-BBBC036 dataset may be due to the relatively blurry images, which complicate the extraction of morphological changes. Based on the MAP metric, PhenoProfiler improves performance by 24.3%, 21.3%, and 15.3% relative to DeepProfiler. In contrast, ResNet50 and ViT exhibit comparable but lower FoE and MAP than PhenoProfiler. Additionally, we compared recall rates at various levels (recall@1, recall@3, recall@5, and recall@10) as shown in **Fig. 2b**. Here the recall rates represent the proportion of biologically related treatments retrieved within the top predictions (**Evaluation Metrics**). Compared to DeepProfiler, using recall@10 as an example, PhenoProfiler achieved improvements of 16.4%, 15.9%, and 9.0% on the BBBC022, CDRP-BIO-BBBC036, and TAORF-BBBC037 datasets, respectively, demonstrating its superior performance in biological matching tasks.

To further illustrate the contributions of each module within PhenoProfiler, we have performed extensive ablation experiments using BBBC022 dataset (**Fig. 2c**). First, we remove the regression learning component within the multi-objective learning module (i.e., "-MSE" option), retaining only classification and contrastive learning. The results show that removing the regression learning results in a notable performance drop, with FoE and MAP decreasing by 16.3% and 5.8%. Next, we test various combinations of loss functions (e.g. "-Con", "-CLS", "-MSE-Con", "-Con-CLS", and "-CLS-MSE"). For example, the removal of both regression and classification learning leads to more performance decrease, with FoE and MAP reduced by 33.7% and 10.9%, respectively. Moreover, we conduct ablation study within the gradient encoder, replacing difference convolution with standard convolution, which results in a reduction of 6% in FoE and 10% in MAP. Additionally, the measurement metrics do not consistently improve as classification loss decreases. As illustrated in **Fig. 2d**, while both MAP and FoE initially increase with decreasing classification loss, they eventually decline. This observation underscores the importance of PhenoProfiler's multi-objective learning design. The optimal weights for multi-objective learning are determined by sensitivity analysis on the BBBC022 datasets. **Fig. 2e** presents the tuned parameters of PhenoProfiler achieved optimal performance (MAP = 0.093, FoE = 52.8). Details are shown in **Parameter Tuning** section.

**PhenoProfiler demonstrates robust generalization and applicability**

To evaluate the generalization of PhenoProfiler, we conduct generalization experiments on the benchmarking datasets using two evaluation strategies: leave-plates-out and leave-dataset-out. In the leave-plates-out strategy, some plates are used as the test set while the remaining plates are used for training. To evaluate cross-dataset generalizability, the leave-dataset-out strategy is used, where one dataset is used for training and the other two serve as test sets. For example, "BBBC022→BBBC036" indicates training on the BBBC022 dataset and validating on BBBC036 (CDRP-BIO-BBBC036).

Regarding the leave-plates-out scenario (**Fig. 3a**), PhenoProfiler consistently surpasses other methods in both FoE and MAP. Specifically, PhenoProfiler has higher FoE than the second-best method by 57.0%, 10.7%, and 13.7%. For the MAP metric, PhenoProfiler outperforms the second-best method by 16.9%, 17.3%, and 6.0%. **Fig. 3b** illustrates the performance comparison in the leave-dataset-out scenario, highlighting the consistent superior performance of PhenoProfiler. For example, in the BBBC022→BBBC036 scenario, PhenoProfiler outperforms the next best method with higher FoE and MAP metrics by 13.0% and 17.1%, respectively. Similarly, in the BBBC022→BBBC037 scenario, PhenoProfiler surpasses the next best method with higher FoE and MAP metrics by 6.9% and 5.4%, respectively. Collectively, PhenoProfiler demonstrates better generalization than existing methods, enhancing more accurate downstream tasks in drug discovery.

To demonstrate the broad applicability and efficacy of PhenoProfiler, we also apply PhenoProfiler to the non-end-to-end scenario. Since the existing three benchmark datasets (BBBC022[22], CDRP-BIO-BBBC036[23], and TAORF-BBBC037[24]) lack cropped single-cell images required for non-end-to-end scenario, here we introduce the Combined Cell Painting data collection (cpg0019[9]). It comprises segmented single-cell images from the three benchmark datasets as well as two additional ones (LUAD-BBBC043 and LINCS), totaling over 8.42 million single-cell images and 450 treatments, meticulously selected to represent a diverse spectrum of phenotypic conditions. Details of this data collection are provided in the **Benchmarking Datasets** section. For non-end-to-end comparisons, we include not only DeepProfiler but also traditional models such as CellProfiler and ImageNet. With the three benchmark datasets as validations, PhenoProfiler exhibits significantly superior performance in non-end-to-end comparisons (**Fig. 3c**). Across the three benchmark datasets as validations, PhenoProfiler achieves 11.9%, 9.7%, and 30.7% higher FoE than the second-best model, DeepProfiler. Similarly, in the MAP metric, PhenoProfiler outperforms the second-best model, ResNet 50, by 10.4%, 8.3%, and 4.7%. These substantial improvements highlight PhenoProfiler's robust capability to capture and represent phenotypic morphology, even in non-end-to-end scenario.

**PhenoProfiler effectively removes batch effects for robust phenotypic representations**

Variations stemming from technical and instrumental factors[9,27] can introduce batch effects between plates subjected to the same treatment, which obscure true biological phenotypic signals and compromise downstream analyses. To evaluate PhenoProfiler's ability to mitigate batch effects, we employed the Inverse Median Absolute Deviation (IMAD) metric, which quantifies the dispersion of image representations. Higher IMAD values indicate reduced dispersion, reflecting successful batch effect correction (details provided in the **Evaluation Metrics** section).

**Fig. 4** illustrates the well-level representation features, with plate IDs distinguished by different colors to highlight batch effects. For DeepProfiler, its representation features extracted from the BBBC022 dataset exhibit clear separation between plate IDs (IMAD = 0.326), indicating the presence of significant plate-specific biases. This separation shows strong technical variations among the extracted representation features, which can obscure true biological signals and hinder downstream analyses. Upon applying extra batch correction step to DeepProfiler's representation features, the UMAP projections displayed improvements in feature cohesiveness (IMAD = 0.458). Notably, the representation features learned by PhenoProfiler exhibit a distinctly more integrated distribution (IMAD = 0.603). This superior performance suggests that PhenoProfiler learns harmonized representation features across different plates, effectively addressing batch effects without additional corrections. This pattern is consistently observed across all three datasets, further validating PhenoProfiler's reliability in generating phenotypic representations that are robust to technical confounders. The inherent capacity of PhenoProfiler to learn harmonized representations directly from raw data not only reduces the need for computationally intensive post-processing but also ensures that biological signals are preserved.

**Phenotype correction strategy of PhenoProfiler improves biological matches**

To effectively capture relative changes under treatments, PhenoProfiler is tailor designed with a phenotype correction strategy (PCs, **Fig. 5a**, details in **Materials and Methods** section) to refine learned phenotypic presentations, which distinguishes PhenoProfiler from existing methods. As illustrated in **Fig. 5a**, PhenoProfiler with PCs corrects image representations by leveraging controlled and treated wells within one plate and emphasizing relative changes in cell phenotypes under treatments.

As shown in **Fig. 5b**, we assess the impact of the PCs through ablation experiments on three benchmark datasets. The results demonstrate that PCs consistently improves the FoE metric while maintaining a nearly unchanged MAP metric (not shown in this panel). We also analyze the aggregation of features before and after incorporating PCs across the three benchmark datasets. UMAP with well-level image representations are quantitatively measured for their dispersion before and after adding PCS using the IMAD metric (**Fig. 5c**). After implementing PCs, the representation features from different plates become significantly more clustered. Additionally, the IMAD metrics increases notably, with substantial improvements of 51.5%, 69.7%, and 11.6% across the three benchmark datasets, respectively.

Furthermore, we analyze the hyperparameter $\alpha$ in PCs (indicated in **Fig. 1c**), which determines the relative weight of controlled and treated wells. As illustrated in **Fig. 5d**, the outer circle of the radar chart corresponds to the value of $\alpha$, with $\alpha = 0$ representing PhenoProfiler without PCs. As $\alpha$ increases from 0 to 1, the relative weight of controlled wells increases, resulting in more pronounced differential effects. Meanwhile, the FoE generally trends upward while the MAP metric remains relatively stable. When $\alpha$ is greater than or equal to 0.7, the FoE begin to reach its maximum. These results demonstrate that PhenoProfiler with PCs can effectively capture the relative changes of cell phenotypes under treatments.

**PhenoProfiler efficiently captures representations of treatment effects**

To visually illustrate the treatment effects, we obtain phenotypic representations using PhenoProfiler under various treatment conditions. **Fig. 6** displays a UMAP projection of PhenoProfiler representations across three benchmark datasets, providing a clear demonstration of how PhenoProfiler captures and organizes biological patterns in both non-end-to-end (**Fig. 6a**) and end-to-end (**Fig. 6b**) scenarios.

For the BBBC022 and CDRP-BIO-BBBC036 datasets, which involve compound treatments, distinct clusters emerge based on their mechanisms of action (MoA). These clusters represent the functional similarities between compounds with shared MoAs, effectively demonstrating how PhenoProfiler translates phenotypic profiles into meaningful groupings that align with known biological functions. Similarly, the TAORF-BBBC037 dataset, which involves gene overexpression perturbations, reveals clear clusters of treatments corresponding to their genetic pathways, such as MAPK and PI3K/AKT. This clustering reaffirms established biological relationships and validates PhenoProfiler's capability to accurately discern pathway-specific features. Notably, these groupings remain consistent across different wells, showcasing PhenoProfiler's robust performance in feature extraction and its reliability in accurately representing treatment effects across experimental variations. However, some clusters are less distinctly discerned in the UMAP projections, suggesting that incorporating additional features, such as compound structures or gene expression data, could enhance the understanding of MoAs. This observation underscores the broader potential of PhenoProfiler to integrate phenotypic, chemical, and gene expression profiles, enabling more comprehensive analyses and gaining deeper mechanistic insights for drug discovery.

## DISCUSSION

In this study, we present PhenoProfiler, an advanced tool for phenotypic representations of cell morphology in drug discovery. PhenoProfiler operates as a fully end-to-end framework, transforming multi-channel images into low-dimensional, biologically meaningful representations. By integrating a multi-objective learning module that incorporates classification, regression, and contrastive learning, PhenoProfiler effectively captures a unified and robust feature space. The extensive benchmarking conducted in this study, involving over 230,000 images from three publicly available datasets, demonstrates that PhenoProfiler significantly outperforms existing state-of-the-art methods in both end-to-end, leave-plates-out, and leave-dataset-out scenarios. Furthermore, in non-end-to-end settings, we utilize over 8.42 million single-cell images for comparative analysis. PhenoProfiler's superior performance on these large-scale datasets across all scenarios highlights its robustness, generalizability, and scalability, positioning it as a highly effective tool for advancing image-based drug discovery.

One challenge in phenotypic representation learning arises from the heterogeneity of cellular responses to treatments. Not all cells exhibit morphological changes upon compound stimulation or perturbation, which can introduce noise into the learned representations if only treatment labels are used. PhenoProfiler addresses this issue by leveraging both morphology image profiles generated by existing methods (e.g., CellProfiler or DeepProfiler) and treatment condition labels as supervisory signals in its multi-objective learning module. The incorporation of pre-extracted features into regression and contrastive learning ensures that PhenoProfiler is better equipped to capture subtle morphological variations while reducing noise in its representations. Given that most cell painting assays are accompanied by features extracted using CellProfiler, these features provide a valuable source of supervisory information to enhance the learning process of PhenoProfiler. In addition, PhenoProfiler employs a phenotype correction strategy that highlights relative phenotypic changes under treatment conditions, facilitating more accurate phenotypic representation learning. This capability makes it an invaluable asset for high-throughput image-based drug screening, offering new opportunities to accelerate the identification of effective therapeutics.

While PhenoProfiler sets a new benchmark in phenotypic representation learning, there are several areas for future explorations and enhancement. First, the design of the multi-objective learning module can be further refined by exploring the interconnections and synergies among classification, regression, and contrastive objectives. Understanding the dependencies between these objectives[28,29] could lead to more cohesive learning strategies. Additionally, while PhenoProfiler currently employs a stepwise training approach to address conflicts between objectives[30,31], future work could focus on developing joint training and optimization techniques to balance these objectives more effectively. Second, recent advancements in large biomedical language models[32–35] offer opportunities to integrate extensive domain knowledge into computational frameworks. Integrating these models' embeddings into PhenoProfiler could enhance its generalizability, robustness, and effectiveness. Last, future efforts should prioritize integrating cell painting assays with complementary data modalities, such as genetic profiles[12,36] and chemical structures[37,38]. Combining these multi-modal data would enable more comprehensive representations, offering a holistic view of cell states and phenotypic responses to various treatments.

The ability of PhenoProfiler to consistently capture and organize complex biological information across diverse datasets and treatment types underscores its versatility and utility in high-throughput drug screening and discovery. By addressing critical challenges in phenotypic profiling, such as scalability, robustness, and interpretability, PhenoProfiler enhances our understanding of treatment effects at the phenotypic level. Additionally, its potential for integrative multi-modal analysis, combining phenotypic data with complementary modalities such as genetic profiles, transcriptomics, and chemical structures, opens new opportunities to explore drug mechanisms and novel drug targets.

**MATERIALS AND METHODS**

**The PhenoProfiler model**

PhenoProfiler is an end-to-end model, with input as multi-channel images and output as phenotypic representations under treatments. The overall framework includes three key modules: a gradient encoder, a transformer encoder, and a multi-objective learning module.

**Gradient encoder**

To enhance the model's ability to understand cell morphology, we design a gradient encoder based on difference convolution $(DC)$[21,22]. This $DC$ function enhances gradient information around cell edges, improving the perception of morphological structures[39,40]. Formally, let $p_0$ represent the central position of a local receptive field $R(p_0)$, with $p_i \in R(p_0)$. The pixel value at position $p_i$ in the input image is denoted as $x_{p_i}$, and the $DC$ function is defined as:

$$DC(x_{p_0}, \theta) = \sum_{p_i \in R(p_0)} w_{p_i} \cdot (x_{p_i} - \theta \cdot x_{p_0}), \quad (1)$$

where $w_{p_i}$ is a learnable parameter, and $\theta \in [0,1]$ is a hyperparameter that controls the balance between semantic and gradient information. When $\theta = 0$, the $DC$ function reduces to traditional convolution ($TC$), i.e. $TC(p_0) = \sum_{p_i \in R(p_0)} w_{p_i} \cdot x_{p_i}$.

The gradient encoder consists of two components: gradient enhancement and residual feature extraction. The gradient enhancement includes two parallel branches of difference convolution: the Deep Gradient branch, denoted as $DG = DC(\cdot, \theta_1)$, and the Shallow Gradient branch, denoted as $SG = DC(\cdot, \theta_2)$. Here $\theta_1 = 0.7$ and $\theta_2 = 0.3$ are used in two branches. The multi-channel input images, represented as $x_{in}$, are processed in parallel through the two branches. The outputs of these branches $H_1$ are then concatenated and passed through a Multi-Layer Perceptron (MLP) layer as follows:

$$G_{DG} = BN\left(\text{ReLU}\left(DG(x_{in})\right)\right), \quad (2)$$

$$G_{SG} = BN\left(\text{ReLU}\left(SG(x_{in})\right)\right), \quad (3)$$

$$H_1 = BN\left(\text{ReLU}\left(\text{MLP}([G_{DG}, G_{SG}])\right)\right). \quad (4)$$

Here $H_1$ is the enhanced latent features. Batch normalization ($BN$) ensures stable and efficient training by normalizing the intermediate latent features.

The subsequent residual feature extraction component employs the ResNet50[15] model, pre-trained on ImageNet[17]. By incorporating residual connections, the ResNet50 model effectively addresses the vanishing gradient problem encountered in training deep neural networks, enabling the network to extract deeper features. This step outputs a fixed-length feature vector $H_2 \in \mathbb{R}^{B \times 2048}$, where $B$ denotes the batch size.

**Transformer encoder**

The latent features $H_2$ is passed through the transformer encoder[16,41] with multi-head self-attention mechanism, effectively capturing global dependencies among features. Then layer normalization and residual connections are applied to ensure stable gradient propagation. Subsequently, a Feed-Forward Neural Network ($FFN$) is used to further extract high-level features, resulting in the output feature vector $H_4 \in \mathbb{R}^{B \times 2048}$. This process is formulated as below:

$$H_3 = MultiHeadAttention(H_2, H_2, H_2), \quad (5)$$

$$H_4 = LayerNorm(H_2 + Dropout(H_3)), \quad (6)$$

$$H_5 = FFN(H_4) = \text{ReLU}(W_2(\text{ReLU}(W_1 H_4 + b_1)) + b_2), \quad (7)$$

$$H_6 = \text{LayerNorm}(H_4 + Dropout(H_5)). \tag{8}$$

Here, *MultiHeadAttention* computes attention weights across multiple attention heads to capture diverse feature dependencies. The *FFN* consists of two fully connected layers with non-linear ReLU activations, parametrized by weights $W_1$, $W_2$, and biases $b_1$, $b_2$. This transformer encoder effectively captures global and local information simultaneously, ensuring the generation of rich and informative feature representations.

**Multi-objective learning**

To facilitate efficient multi-objective learning, we use a feature projection that maps latent features to a lower-dimensional space, facilitating efficient multi-objective learning. Specifically, the latent feature $H_6 \in \mathbb{R}^{B \times 2048}$ is first linearly transformed to a lower-dimensional space. A GELU activation function is then applied to introduce non-linear characteristics, followed by further processing through a fully connected layer with dropout to prevent overfitting. This feature projection process provides final output $\hat{Z} \in \mathbb{R}^{B \times 672}$ as follows:

$$Z_1 = W_3 H_6 + b_3, \tag{9}$$

$$Z_2 = \text{Dropout}(W_4 \cdot \text{GELU}(Z_1) + b_4), \tag{10}$$

$$\hat{Z} = \text{LayerNorm}(Z_1 + Z_2). \tag{11}$$

Then a classification head is implemented as a simple linear layer maps output representations $\hat{Z}$ to $\hat{y}$, representing the predicted treatment labels.

*Classification learning.* To characterize the phenotypic responses of cells under various treatments, we utilize cross-entropy loss to evaluate the discrepancy between the model's predictions and ground truth:

$$\mathcal{L}_{CLS} = -\sum_{i=1}^{N} y_i \log(\hat{y}_i). \tag{12}$$

Here, $N$ represents the number of treatment categories, and $y_i$ is the one-hot encoding of the treatment label in ground truth. If the ground truth is treatment category $i$, then $y_i = 1$; otherwise, $y_i = 0$.

*Regression learning.* Here we have innovatively designed a regression learning to learn cell morphological representations. Compared to classification labels, regression learning provides richer and continuous feature information, enabling the model to capture subtle morphological changes under different treatments, thus more accurately reflecting cell morphology. To obtain regression labels, we use the median of the morphology profiles of cpg0019 as the regression morphology labels and train the model using Mean Squared Error (MSE) loss:

$$\mathcal{L}_{MSE} = \frac{1}{m} \sum_{i=1}^{M} (Z_i - \hat{Z}_i)^2, \tag{13}$$

where $Z_i$ is the morphology profiles of the $i$-th image, $\hat{Z}_i$ is predicted morphology features, and $M$ is the number of images.

*Contrastive learning.* Contrastive learning enhances the model 's ability to distinguish features by maximizing the similarity between similar images and minimizing the similarity between dissimilar ones. Specifically, contrastive learning does not rely on ground truth labels but focuses on learning feature representations based on the relative relationships between images. This approach not only reduces the negative impact of noisy labels but also improves the model's robustness and generalization when handling unseen data. The contrastive loss function is formulated as follows:

$$\mathcal{L}_{Con} = -\frac{1}{B} \sum_{i=1}^{B} \log \exp(\hat{Z}_i \cdot Z_i / \tau) / \sum_{j=1}^{B} \exp(\hat{Z}_i \cdot Z_i / \tau), \tag{14}$$

where $B$ is the batch size, $\hat{Z}_i$ denotes the predicted morphology features of the $i$-th image, and $Z_i$ is the morphology profiles, and $\tau$ is the temperature parameter. This objective trains the model to produce discriminative feature vectors by maximizing the similarity between matching image-embedding pairs while minimizing the similarity between non-matching pairs.

*Multi-objective loss.* By integrating classification, regression, and contrastive learning, PhenoProfiler provides a unified and robust feature space, comprehensively learns the image representations of multi-channel cell images. This multi-object learning architecture enhances the overall generalization performance of the model. To achieve an effective balance in multi-object learning, we assign a weight parameter to each object and adjust these weight parameters to balance the losses. The final total loss function can be expressed as:

$$\mathcal{L}_{\text{total}} = \lambda_1 \mathcal{L}_{\text{CLS}} + \lambda_2 \mathcal{L}_{\text{MSE}} + \lambda_3 \mathcal{L}_{\text{Con}}, \tag{15}$$

where $\lambda_1, \lambda_2, \lambda_3$ are the weight parameters for the classification, regression, and contrastive learning, respectively. Based on extensive ablation experiments (see **Fig. 2e**), we set these weights to 0.1, 100, and 1, respectively. This multi-loss balancing strategy enables PhenoProfiler to find the optimal trade-off among different objects, thereby enhancing the overall performance and robustness of the model. Additionally, this strategy allows us to flexibly adjust the weights of each learning according to the specific requirements of the application scenario, achieving optimal feature representation and prediction performance.

**Parameter Tuning**

PhenoProfiler is an end-to-end multi-channel image encoder designed to convert multi-channel images into corresponding morphology representations. During training, conflicts among objects in multi-objective joint training caused the model to struggle to converge to an optimal state[30,31]. To address this, we adopted a stepwise training strategy. Initially, we trained the regression learning using MSE loss to optimize the model. After approximately 100 epochs, we proceeded with joint optimization based on the multi-object learning architecture. The hyperparameter settings were as follows: a batch size of 300, a maximum of 200 training epochs, and 12 workers. The learning rate followed a staged decay strategy: 2e-3 for the first 10 epochs, 1e-3 for the next 50 epochs, 5e-4 for the subsequent 60 epochs, and 1e-4 for the final 80 epochs. The training environment was Ubuntu 22.04, utilizing four NVIDIA A100 GPUs (40GB version).

Based on findings from single classification learning in **Fig. 2d**, we first fixed the classification weight at $\lambda_1 = 1$ and then explored the performance relationship with varying weights for regression ($\lambda_2$) and contrastive ($\lambda_3$) learning. We investigated combinations centered around 100, spanning 10 times on either side. The results indicate that as $\lambda_2$ increases, model performance gradually improves, underscoring the significant impact of regression loss. Notably, when $\lambda_2 = 1000$ and $\lambda_3 = 10$, the model achieved optimal performance (MAP = 0.093, FoE = 52.8). To visually represent performance changes under different conditions, we created a line chart illustrating the sensitivity of the multi-object learning design. As shown in the bottom right corner of **Fig. 2e**, the five lines correspond to different values of $\lambda_2$, with the x-axis representing $\lambda_3$. It is evident that with $\lambda_2 = 1000$, the model consistently performs well, highlighting the substantial contribution of regression learning and the advantages of the multi-object learning architecture.

**Model Inference**

In model inference, we first clarify the four levels of data involved in this task. The dataset comprises four levels of features: plate-level, treatment-level, well-level, and site-level. Specifically, a dataset contains $P$ plates, each plate includes $T$ treatments, each treatment corresponds to $W$ wells, and each well contains $S$ sites, with each site corresponds to a multi-channel image. Hence, the output of PhenoProfiler is a site-level feature $\widehat{\mathbf{Z}}$. The values of $P$, $T$, $W$, and $S$ vary across datasets. Following the validation procedure used in DeepProfiler, we applied mean aggregation at both stages to get the next level aggregated features. After obtaining well-level features, we employed Sphering transform as a batch correction method to minimize confounders. Through these aggregation processes, we obtained the final treatment-level features for evaluation.

Phenotype Correction Strategy (PCs) directly optimizes the output of the PhenoProfiler in a plate. It aims to leverage the differences between treated wells and controlled wells within a plate to refine image representation. The implementation process of PCs is detailed in **Fig. 1c**. First, we calculate the mean of all control wells in the current plate, denoted as $\mathbf{W}^i$, where $i$ represents the $i$-th plate. Assuming that a plate contains $C$ controlled wells, the calculation is as follows:

$$\mathbf{W}^i = \frac{1}{C * S} \sum_{j=1}^{C} \sum_{k=1}^{S} \widehat{\mathbf{Z}}_{jk}. \tag{16}$$

Then, define $\alpha$ as a hyperparameter that balances the weight between controlled wells and treated wells, $\widehat{E}_{ij}$ represent the predicted embedding vector for the $j$-th well in the $i$-th plate. $\alpha * W^i$ is subtracted from all predicted embeddings $\widehat{E}_{ij}$ in the current plate to obtain the refined embeddings $E_{ij}$. The process is as follows:

$$E_{ij} = \widehat{E}_{ij} - \alpha * W^i. \tag{17}$$

PCs serves as a correction and optimization operation applied to the extracted features, making it virtually cost-free and plug-and-play.

**Benchmarking Methods**

To intuitively evaluate PhenoProfiler's performance within an end-to-end pipeline, we compared it against three baseline methods: DeepProfiler[9], ResNet50[15], and ViT-Base[16]. We made uniform adjustments to apply these models to multi-channel image processing. Specifically, to handle five-channel cell painting images, we included a convolutional layer at the front of each model to adjust the number of input image channels from five to three and reduce the dimensionality. The networks were initialized using the corresponding pre-trained weights provided by the "timm" library. For example, the ViT-Base model utilized the weights for vit_base_patch32_224. Notably, we adopted a fully trainable parameter mode for all models, without freezing any task weights, and maintained consistent training methods and hyperparameters throughout the training process.

In the non-end-to-end scenarios, we included two additional comparison models: CellProfiler[18] and ImageNet[17]. CellProfiler is an open-source software tool designed for measuring and analyzing cell images, providing a robust platform for extracting quantitative data from biological images. It enables researchers to identify and quantify phenotypic changes effectively. ImageNet, following the design outlined in DeepProfiler, involves pre-training models on the large-scale ImageNet dataset. By incorporating these models, we aimed to benchmark our pipeline's performance against established standards in the field, ensuring a comprehensive evaluation of its capabilities.

**Benchmarking Datasets**

The PhenoProfiler model leverages five distinct datasets: BBBC022[24], CDRP-BIO-BBBC036[25], TAORF-BBBC037[26], LUAD-BBBC043[42], and LINCS[43]. Among these, BBBC022, CDRP-BIO-BBBC036, and LINCS focus on phenotypic responses to compound treatments, whereas TAORF-BBBC037 and LUAD-BBBC043 address phenotypic responses to gene overexpression using open reading frames (ORFs). Together, these datasets form a comprehensive dataset of over 230,000 multi-channel images. These images encompass two treatment types (compounds and gene overexpression), two control types (empty and DMSO), and two cell lines (A549 and U2OS), collected from 231 plates. Detailed dataset information is summarized in **Supplementary Table 1**. To facilitate data storage and transmission, we apply image compression and illumination correction to convert the images from TIFF to PNG format, achieving approximately six times the compression without significant quality loss[9]. To further validate the cell morphology representation capability of PhenoProfiler in a non-end-to-end pipeline, we introduce the cpg0019 dataset[9]. This dataset comprises 8.4 million single-cell images, which were cropped from multi-channel images across these five datasets. It includes 450 treatments, meticulously selected to represent a diverse array of phenotypic responses.

Due to the different input data requirements for end-to-end and non-end-to-end scenarios, we have trained and validated them on different datasets. For the end-to-end scenario, we directly used the BBBC022, CDRP-BIO-BBBC036, and TAORF-BBBC037 datasets for training and validation. For the non-end-to-end scenario, we segment cells from the aforementioned five datasets, splitting them into single-cell sub-images. A subset of these sub-images is combined to form the cpg0019 dataset. It is important to note that since the other two datasets do not contain ground truth, we test both pipelines using the BBBC022, CDRP-BIO-BBBC036, and TAORF-BBBC037 datasets, with the difference being whether the input images are cropped.

Given the distinct input data requirements for end-to-end and non-end-to-end pipelines, we train and validate them using different datasets. For the end-to-end pipeline, we directly utilize the BBBC022, CDRP-BIO-BBBC036, and TAORF-BBBC037 datasets. Conversely, for the non-end-to-end pipeline, we segment cells from the aforementioned five datasets into single-cell sub-images, combining a subset to form the cpg0019 dataset. Notably, since the other two datasets lack ground truth, we tested both pipelines using the BBBC022, CDRP-BIO-BBBC036, and TAORF-BBBC037 datasets, with the primary difference being whether the input images are cropped.

Data preprocessing involves three main components: (1) Image data preprocessing: We stack images from different channels in the order of ['DNA', 'ER', 'RNA', 'AGP', 'Mito'] to obtain multi-channel images. The images are resized to a uniform size of (5, 448, 448) pixels and the pixel values are scaled to the range of 0 to 1. (2) Classification label preprocessing: We read the CSV file and initialize the label encoder. Labels are encoded based on the column names (Treatment or pert_name) in the CSV file. (3) Morphology profiles preprocessing for the regression and contrastive learning. For the supervisory labels in regression and contrastive learning, we select the median values of the morphology profiles provided in the cpg0019 dataset.

**Evaluation Metrics**

We evaluate the model's capacity to represent cell morphology using a reference collection of treatments to identify biological matches in treatment experiments. Following strategies outlined in previous studies[9,23,44,45], we implement a biological matching task where users can search for treatments associated with the same MoA or genetic pathway, applicable to both compound and gene overexpression perturbations. Initially, we aggregate features from various methods at the treatment level and assess the relationships among these treatments within the feature space, guided by established biological connections. This approach allows us to identify treatments that are proximally situated, suggesting potential similarities in their biological effects or mechanisms of action. This evaluation not only demonstrates the model's effectiveness in representing cell morphology but also offers a valuable framework for advancing biological research. To quantify the similarity between query treatments, we utilize cosine similarity and generate a ranked treatment list based on relevance, presented in descending order. A positive result is achieved if at least one biological annotation in the sorted list matches the query; otherwise, the result is regarded as negative.

For evaluating the quality of results for a given query, we employ two primary metrics: 1) Folds of Enrichment (FoE) and 2) Mean Average Precision (MAP).

1) Folds of Enrichment (FoE): This metric assesses the overrepresentation of predicted features in the reference set, indicating the model's ability to identify relevant biological treatments. We calculate the odds ratio using a one-sided Fisher's exact test for each query treatment, which employs a 2×2 contingency table. The first row contains the counts of treatments with the same MoAs or pathways (positive matches) versus those with different MoAs or pathways (negative matches) above a pre-defined threshold. The second row contains the corresponding counts for treatments below the threshold. The odds ratio is computed as the sum of the first row divided by the sum of the second row, estimating the likelihood of observing treatments sharing the same MoA or pathway among the top connections. We average the odds ratios across all query treatments, with the threshold set at the top 1% of connections, thus anticipating significant enrichment for positive matches[23].

2) Mean Average Precision (MAP): For each query treatment, we calculate the average precision, which is the area under the precision-recall curve, following standard practices in information retrieval. The evaluation process starts with the result most similar to the query and continues until all relevant pairs (those with the same mechanism of action or pathway) are identified. MAP effectively captures both the precision and recall of the model's predictions, offering insights into its reliability and robustness. The specific calculation process is as follows:

$$Precision = \frac{TP}{TP + FP}, \tag{18}$$

$$Recall = \frac{TP}{TP + FN}. \tag{19}$$

Here, *TP* stands for true positives, *FP* for false positives, and *FN* for false negatives. MAP refers to the average precision across multiple queries. For each query, we calculate the area under the precision-recall curve and then average these values across all queries. Given that the number of MoAs or pathways varies, precision and recall are interpolated for each query to cover the maximum number of recall points. The interpolated precision at each recall point is:

$$P_{inter}(r) = \max_{r' \geq r} P(r').  \tag{20}$$

The average precision of a query treatment is the mean of the interpolated precision values $P_{inter}$ at all recall points. The reported MAP is the average of the average precision values across all queries. Together, these metrics provide a rigorous assessment of PhenoProfiler's performance in capturing biological relevance in treatment experiments and underscore its utility in phenotypic drug discovery.

We also used recall rates at various levels (recall@1, recall@3, recall@5, and recall@10) to evaluate the model's performance. Recall@K is an important metric in information retrieval and recommendation systems, indicating the proportion of correct results within the top K returned results. For instance, recall@5 represents the proportion of biologically related treatments retrieved within the top five positions of a predicted treatments ranked list. This is crucial for user experience, as users typically only look at the first few results, and the relevance of these results directly impacts user satisfaction.

In addition to the main metrics mentioned above, we introduced the Inverse Median Absolute Deviation (IMAD) metric to quantitatively evaluate the aggregation degree of features. The calculation steps for the IMAD metric are as follows: First, Principal Component Analysis (PCA) is performed to reduce the dimensionality of the data, retaining 95% of the variance. Then, we use Uniform Manifold Approximation and Projection (UMAP) for further dimensionality reduction and embedding. Next, we combine the UMAP 1 and UMAP 2 columns into a coordinate array and calculate the pairwise distances between all points. Subsequently, we compute the Median Absolute Deviation (MAD) of these distances. Finally, we take the reciprocal of the MAD to obtain the IMAD. A higher IMAD indicates a tighter aggregation of the data.

**FIGURE LEGENDS**

**Fig. 1: Framework of the PhenoProfiler for morphology representations**. **a** Flowchart comparison of end-to-end PhenoProfiler with existing non-end-to-end methods. **b** PhenoProfiler includes a gradient encoder to enhance edge gradients, improving clarity and contrast in cell morphology. A transformer encoder then captures high-dimensional dependencies and intricate relationships, enriching image representations. A designed multi-objective learning module is utilized for accurate morphological representation learning. **c** For model inference, PhenoProfiler uses phenotype correction strategy (PCs) with hyperparameter α to identify morphological changes between treated and control conditions.

**Fig. 2: Performance analysis of benchmarking methods in biological matching tasks**. **a** Comparison of end-to-end feature representation performance across different methods in biological matching tasks using three benchmark datasets (BBBC022, CDRP-BIO-BBBC036, and TAORF-BBBC037), evaluated with two evaluation metrics (MAP, FoE), and three comparison methods (DeepProfiler, ResNet50, ViT). **b** Performance comparison of different methods at different recall rates (recall@1, recall@3, recall@5, and recall@10). **c** Ablation experiments of PhenoProfiler, showing performance changes after sequential removal of each module. Specifically, "-MSE", "-Con", and "-CLS" represent the removal of regression, contrastive, and classification learning in the multi-objective module, while "-Gradient" represents the exclusion of difference operations. **d** Performance curve of PhenoProfiler under solely classification learning, showing variations in MAP and FoE as the classification loss decreases. **e** Sensitivity analysis of multi-objective learning of PhenoProfiler, exploring the impact of regression and contrastive learning ($\lambda_2$ and $\lambda_3$) while maintaining fixed classification learning. MAP: Mean Average Precision; FoE: Folds of Enrichment; MSE: Mean Squared Error; Con: Contrastive Learning; CLS: Classification Learning.

**Fig. 3: Benchmarking zero-shot performance across multiple datasets. a** Performance evaluation of different models using leave-plates-out validation. **b** Performance evaluation of different models using leave-dataset-out validation. **c** Comparison of feature representation performance of multiple models (DeepProfiler, CellProfiler, ImageNet, ResNet50, ViT) in non-end-to-end applications. Validations are performed on three benchmark datasets (BBBC022, BBBC036, and BBBC037). FoE: Folds of Enrichment; MAP: Mean Average Precision; BBBC036: CDRP-BIO-BBBC036; BBBC037: TAORF-BBBC037.

**Fig. 4: Robustness of feature representations across different methods.** UMAP visualizations of well-level features predicted by DeepProfiler and PhenoProfiler across three benchmark datasets. Features are colored by plate IDs and treatment conditions (control vs. treatment). IMAD quantifies the cohesiveness of the UMAP patterns, with higher values indicating better batch effects removal. UMAP: Uniform Manifold Approximation and Projection; BC: Batch Correction; IMAD: Inverse Median Absolute Deviation.

**Fig. 5: Quantitative analysis of the phenotype correction strategy in PhenoProfiler.** **a** The conceptual motivation for the design of PCs. Phenotypic differences between treated- and controlled- wells capture the treatment response. **b** Ablation experiments demonstrating the impact of PCs across three datasets, showing a consistent increase in the FoE and MAP with the inclusion of PCs. **c** UMAP visualizations of feature representations generated by PhenoProfiler, with and without PCs. **d** Sensitivity analysis of the hyperparameter α in the PCs, evaluating the harmonization of well-level features using IMAD. PCs: Phenotype correction strategy; IMAD: Inverse Median Absolute Deviation; FoE: Folds of Enrichment; MAP: Mean Average Precision; UMAP: Uniform Manifold Approximation and Projection.

**Fig. 6: Quantitative and qualitative evaluation of feature representations of treatment effects. a** UMAP projections of treatment profiles using well-level features provided by PhenoProfiler in non-end-to-end scenario. **b** UMAP projections of treatment profiles using well-level features provided by PhenoProfiler in end-to-end scenario. Well-level profiles, control wells, and treatment-level profiles are included. Text annotations highlight clusters where all or most points share same biological annotations for treatment-level profiles. UMAP: Uniform Manifold Approximation and Projection.

**Supplementary Table 1. Benchmarking datasets for model performance evaluation.**

# DATA AVAILABILITY

All experiments in this study utilized publicly available datasets, which can be accessible from public S3 buckets. To download the data, you need to install the AWS CLI that matches your device by following the instructions at [AWS CLI Installation Guide](). Use the following command with the cp or sync command, along with the –recursive and --no-sign-request flags for data retrieval. For the BBBC022 dataset can be downloaded with the following command: "aws s3 cp s3://cytodata/datasets/Bioactives-BBBC022-Gustafsdottir/ ./ --recursive --no-sign-request". For the other datasets, use the following commands: CDRP-BIO-BBBC036: "aws s3 cp s3://cytodata/datasets/CDRPBIO-BBBC036-Bray/ ./ --recursive --no-sign-request"; TAORF-BBBC037: "aws s3 cp s3://cytodata/datasets/TA-ORF-BBBC037-Rohban/ ./ --recursive --no-sign-request". For the non-end-to-end dataset cpg0019: "aws s3 cp s3://cellpainting-gallery/cpg0019-moshkov-deepprofiler/ ./ --recursive --no-sign-request".

# CODE AVAILABILITY

All source codes and trained models in our experiments have been made publicly available at https://github.com/QSong-github/PhenoProfiler.

## Web Server Implementation

The PhenoProfiler web server integrates a JavaScript/TypeScript frontend (Next.js framework) with server-side rendering and Tailwind CSS styling, enhanced with React hooks for dynamic interactivity. The Python Django backend supports scalable data processing, coupled with an SQLite database for lightweight storage. A RESTful API coordinates client-server communication, enabling seamless file uploads and standardized JSON responses. Deployment utilizes the Caddy server with automated HTTPS/SSL and reverse proxy configurations, ensuring secure and efficient operations. The webserver is publicly accessible at https://phenoprofiler.org and does not require coding skills.

# ACKNOWLEDGMENTS

B. Z. was supported by the University of Macau under Grant MYRG-GRG2024-00205-FST. This work partially used Jetstream2[46] through allocation CIS230237 from the Advanced Cyberinfrastructure Coordination Ecosystem: Services & Support (ACCESS)[47] program, which is supported by National Science Foundation grants #2138259, #2138286, #2138307, #2137603, and #2138296. The authors thank Srinivas Niranj Chandrasekaran, Postdoctoral Associate at the Imaging Platform, Broad Institute of MIT and Harvard, for his guidance, discussions, and support throughout this project.

# REFERENCES

1. Vincent, F. *et al.* Phenotypic drug discovery: recent successes, lessons learned and new directions. *Nat. Rev. Drug Discov.* **21**, 899–914 (2022).

2. Chandrasekaran, S. N., Ceulemans, H., Boyd, J. D. & Carpenter, A. E. Image-based profiling for drug discovery: due for a machine-learning upgrade? *Nat. Rev. Drug Discov.* **20**, 145–159 (2021).

3. Seal, S. *et al.* Cell Painting: a decade of discovery and innovation in cellular imaging. *Nat. Methods* (2024) doi:10.1038/s41592-024-02528-8.

4. Cross-Zamirski, J. O. *et al.* Label-free prediction of cell painting from brightfield images. *Sci. Rep.* **12**, 10001 (2022).

5. Chandrasekaran, S. N. *et al.* JUMP Cell Painting dataset: morphological impact of 136,000 chemical and genetic perturbations. Preprint at https://doi.org/10.1101/2023.03.23.534023 (2023).

6. Chandrasekaran, S. N. *et al.* Three million images and morphological profiles of cells treated with matched chemical and genetic perturbations. *Nat. Methods* **21**, 1114–1121 (2024).

7. Haghighi, M., Caicedo, J. C., Cimini, B. A., Carpenter, A. E. & Singh, S. High-dimensional gene expression and morphology profiles of cells across 28,000 genetic and chemical perturbations. *Nat. Methods* **19**, 1550–1557 (2022).

8. Bray, M.-A. *et al.* Cell Painting, a high-content image-based assay for morphological profiling using multiplexed fluorescent dyes. *Nat. Protoc.* **11**, 1757–1774 (2016).

9. Moshkov, N. *et al.* Learning representations for image-based profiling of perturbations. *Nat. Commun.* **15**, 1594 (2024).

10. Tian, G., Harrison, P. J., Sreenivasan, A. P., Carreras-Puigvert, J. & Spjuth, O. Combining molecular and cell painting image data for mechanism of action prediction. *Artif. Intell. Life Sci.* **3**, 100060 (2023).

11. Morphological cell profiling of SARS-CoV-2 infection identifies drug repurposing candidates for COVID-19. https://www.pnas.org/doi/epub/10.1073/pnas.2105815118 doi:10.1073/pnas.2105815118.

12. Way, G. P. *et al.* Morphology and gene expression profiling provide complementary information for mapping cell state. *Cell Syst.* **13**, 911-923.e9 (2022).

13. Shihang Wang, Qilei Han, Weichen Qin, Lin Wang, Junhong Yuan, Yiqun Zhao, Pengxuan Ren, Yunze Zhang, Yilin Tang, Ruifeng Li, Zongquan Li, Wenchao Zhang, Shenghua Gao, Fang Bai. PhenoScreen: A Dual-Space Contrastive Learning Framework-based Phenotypic Screening Method by Linking Chemical Perturbations to Cellular Morphology. *bioRxiv* (2024) doi:https://doi.org/10.1101/2024.10.23.619752.

14. Caicedo, J. C. *et al.* Data-analysis strategies for image-based cell profiling. *Nat. Methods* **14**, 849–863 (2017).

15. He, K., Zhang, X., Ren, S. & Sun, J. Deep Residual Learning for Image Recognition. in *2016 IEEE Conference on Computer Vision and Pattern Recognition (CVPR)* 770–778 (IEEE, Las Vegas, NV, USA, 2016). doi:10.1109/CVPR.2016.90.

16. Dosovitskiy, A. *et al.* An Image is Worth 16x16 Words: Transformers for Image Recognition at Scale. Preprint at http://arxiv.org/abs/2010.11929 (2021).

17. Deng, J. *et al.* ImageNet: A Large-Scale Hierarchical Image Database.

18. McQuin, C. *et al.* CellProfiler 3.0: Next-generation image processing for biology. *PLOS Biol.* **16**, e2005970 (2018).

19. Tan, M. & Le, Q. EfficientNet: Rethinking Model Scaling for Convolutional Neural Networks. in *Proceedings of the 36th International Conference on Machine Learning* 6105–6114 (PMLR, 2019).

20. Bao, Y., Sivanandan, S. & Karaletsos, T. Channel Vision Transformers: An Image Is Worth 1 x 16 x 16 Words. Preprint at http://arxiv.org/abs/2309.16108 (2024).

21. Yu, Z. *et al.* Searching Central Difference Convolutional Networks for Face Anti-Spoofing. in *2020 IEEE/CVF Conference on Computer Vision and Pattern Recognition (CVPR)* 5294–5304 (IEEE, Seattle, WA, USA, 2020). doi:10.1109/CVPR42600.2020.00534.

22. Yu, Z., Qin, Y., Zhao, H., Li, X. & Zhao, G. Dual-Cross Central Difference Network for Face Anti-Spoofing. Preprint at http://arxiv.org/abs/2105.01290 (2021).

23. Rohban, M. H., Abbasi, H. S., Singh, S. & Carpenter, A. E. Capturing single-cell heterogeneity via data fusion improves image-based profiling. *Nat. Commun.* **10**, 2082 (2019).

24. Gustafsdottir, S. M. *et al.* Multiplex Cytological Profiling Assay to Measure Diverse Cellular States. *PLoS ONE* **8**,

e80999 (2013).

25. Bray, M.-A. *et al.* A dataset of images and morphological profiles of 30 000 small-molecule treatments using the Cell Painting assay. *GigaScience* **6**, giw014 (2017).

26. Rohban, M. H. *et al.* Systematic morphological profiling of human gene and allele function via Cell Painting. *eLife* **6**, e24060 (2017).

27. Arevalo, J. *et al.* Evaluating batch correction methods for image-based cell profiling. *Nat. Commun.* **15**, 6516 (2024).

28. Chen, S., Zhang, Y. & Yang, Q. Multi-Task Learning in Natural Language Processing: An Overview. *ACM Comput. Surv.* **56**, 1–32 (2024).

29. Allenspach, S., Hiss, J. A. & Schneider, G. Neural multi-task learning in drug design. *Nat. Mach. Intell.* **6**, 124–137 (2024).

30. Gong, Z. *et al.* CoBa: Convergence Balancer for Multitask Finetuning of Large Language Models. Preprint at http://arxiv.org/abs/2410.06741 (2024).

31. Tiomoko, M., Tiomoko, H. & Couillet, R. Deciphering and Optimizing Multi-Task Learning: a Random Matrix Approach. in *ICLR 2021 - 9th International Conference on Learning Representations* (Vienna, Austria, 2021).

32. Wang, B. *et al.* Pre-trained Language Models in Biomedical Domain: A Systematic Survey. *ACM Comput. Surv.* **56**, 1–52 (2024).

33. Ma, T. *et al.* Y-Mol: A Multiscale Biomedical Knowledge-Guided Large Language Model for Drug Development. Preprint at http://arxiv.org/abs/2410.11550 (2024).

34. Cui, H. *et al.* scGPT: toward building a foundation model for single-cell multi-omics using generative AI. *Nat. Methods* **21**, 1470–1480 (2024).

35. Hao, M. *et al.* Large-scale foundation model on single-cell transcriptomics. *Nat. Methods* **21**, 1481–1491 (2024).

36. Iyer, N. S. *et al.* Cell morphological representations of genes enhance prediction of drug targets. *bioRxiv* 2024–06 (2024).

37. Sanchez-Fernandez, A., Rumetshofer, E., Hochreiter, S. & Klambauer, G. CLOOME: contrastive learning unlocks

bioimaging databases for queries with chemical structures. *Nat. Commun.* **14**, 7339 (2023).

38. Zheng, S. *et al.* SYSU-Cross-Modal Graph Contrastive Learning with Cellular Images. *Adv. Sci.* **11**, 2404845 (2024).

39. Li, B. *et al.* Gene expression prediction from histology images via hypergraph neural networks. *Brief. Bioinform.* **25**, bbae500 (2024).

40. Li, B. *et al.* Exponential distance transform maps for cell localization. *Eng. Appl. Artif. Intell.* **132**, 107948 (2024).

41. Chen, H. *et al.* Pre-Trained Image Processing Transformer. in *2021 IEEE/CVF Conference on Computer Vision and Pattern Recognition (CVPR)* 12294–12305 (IEEE, Nashville, TN, USA, 2021). doi:10.1109/CVPR46437.2021.01212.

42. Caicedo, J. C. *et al.* Cell Painting predicts impact of lung cancer variants. *Mol. Biol. Cell* **33**, ar49 (2022).

43. Way, G. P. *et al.* Predicting cell health phenotypes using image-based morphology profiling. *Mol. Biol. Cell* **32**, 995–1005 (2021).

44. Wolf, T. *et al.* Transformers: State-of-the-Art Natural Language Processing. in *Proceedings of the 2020 Conference on Empirical Methods in Natural Language Processing: System Demonstrations* (eds. Liu, Q. & Schlangen, D.) 38–45 (Association for Computational Linguistics, Online, 2020). doi:10.18653/v1/2020.emnlp-demos.6.

45. Ljosa, V. *et al.* Comparison of Methods for Image-Based Profiling of Cellular Morphological Responses to Small-Molecule Treatment. *J. Biomol. Screen.* **18**, 1321–1329 (2013).

46. Hancock, D.Y. et al. Practice and Experience in Advanced Research Computing 2021: Evolution Across All Dimensions. in (Association for Computing Machinery, Boston, MA, USA).

47. Boerner, T.J., Deems, S., Furlani, T.R., Knuth, S.L., Towns, J. Practice and Experience in Advanced Research Computing 2023: Computing for the Common Good. in 173–176 (Association for Computing Machinery, Portland, OR, USA).

**Fig. 1**

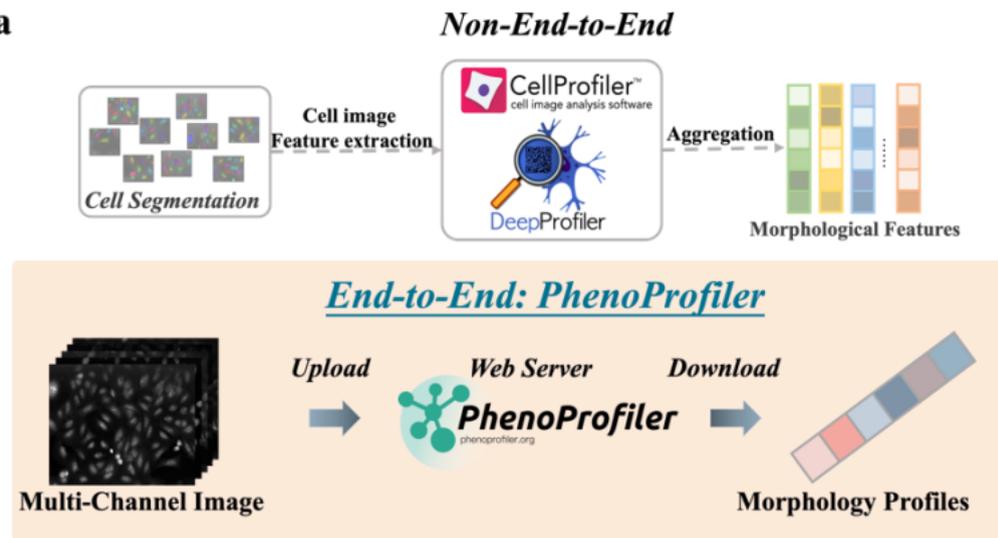
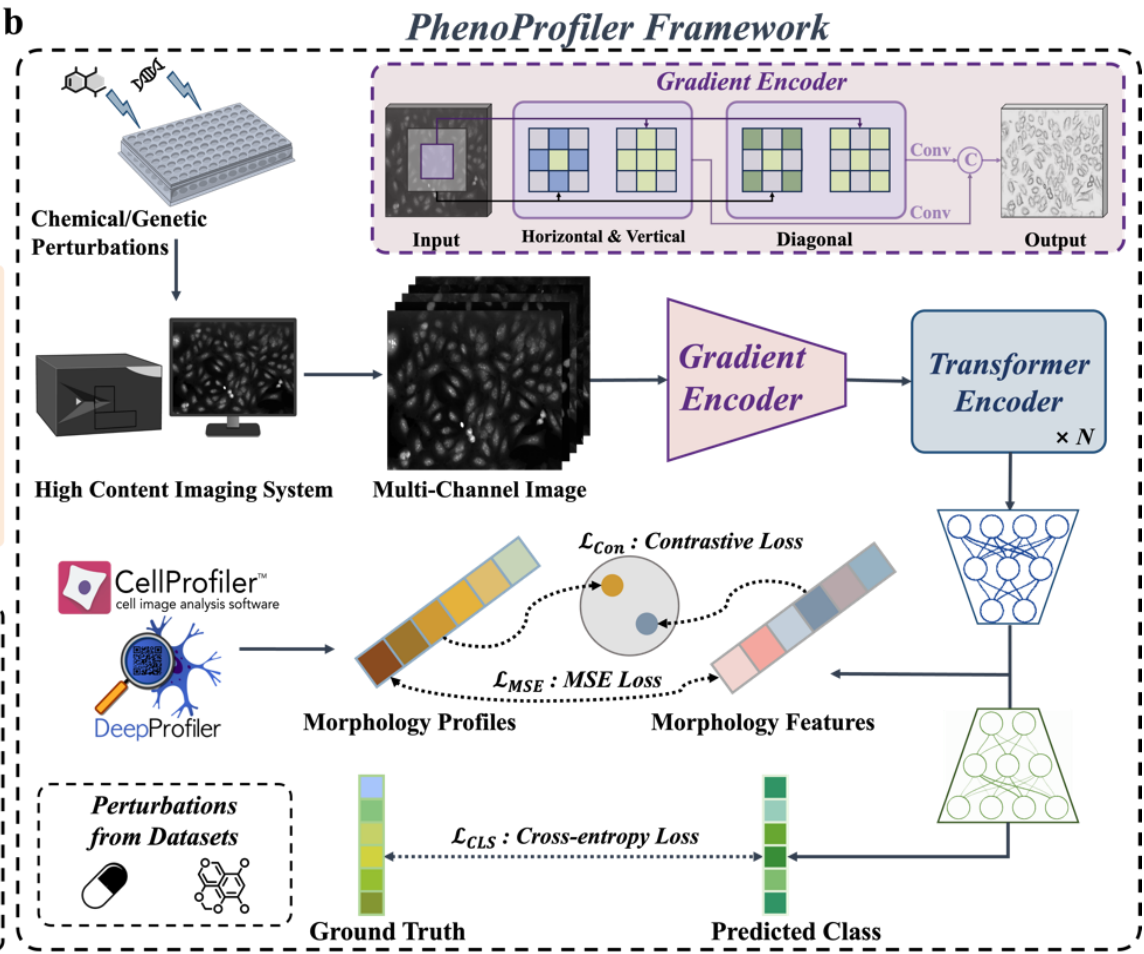
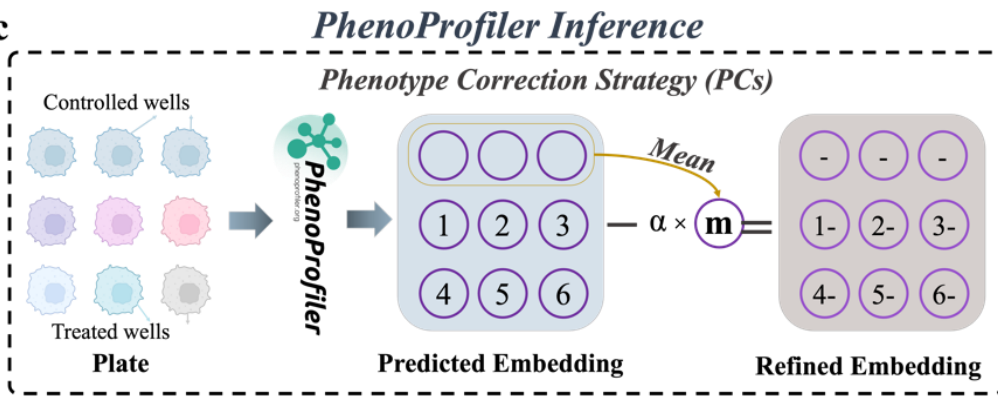

**Fig. 2**

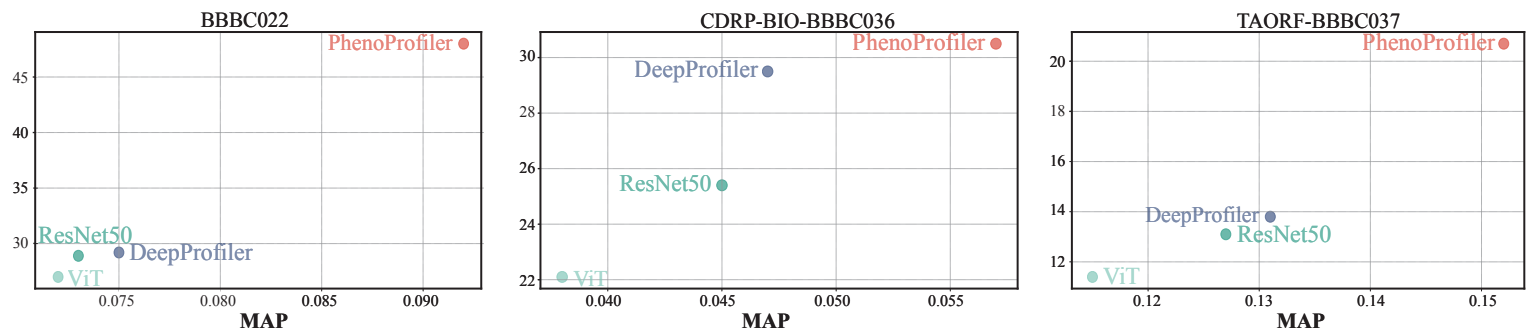

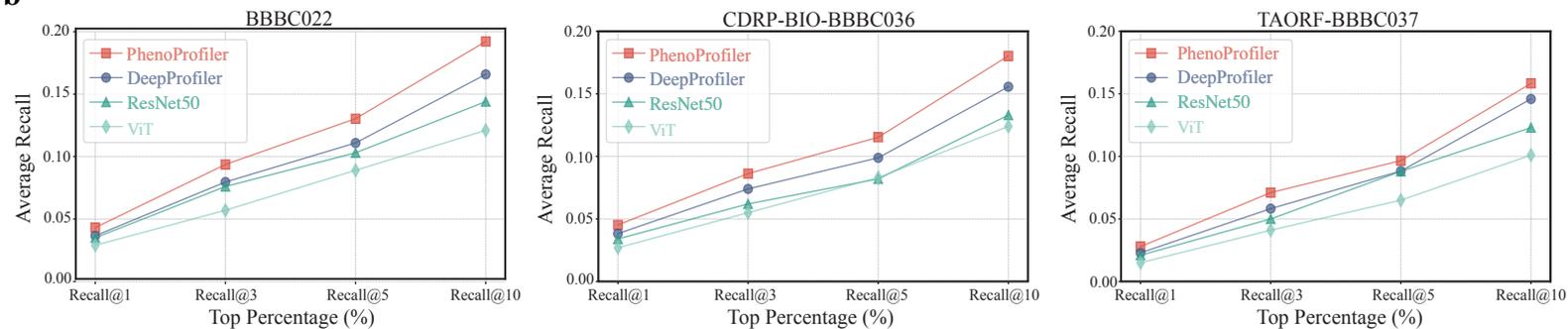

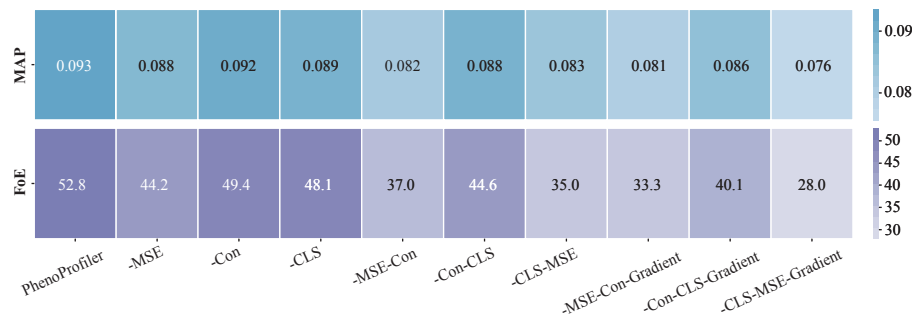

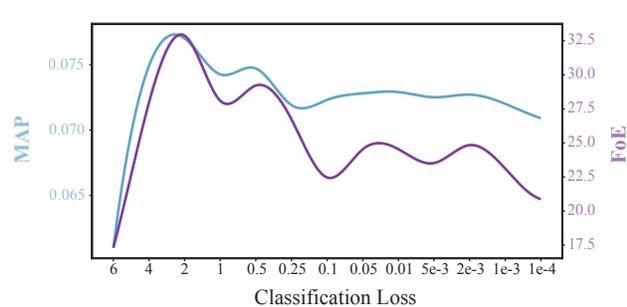

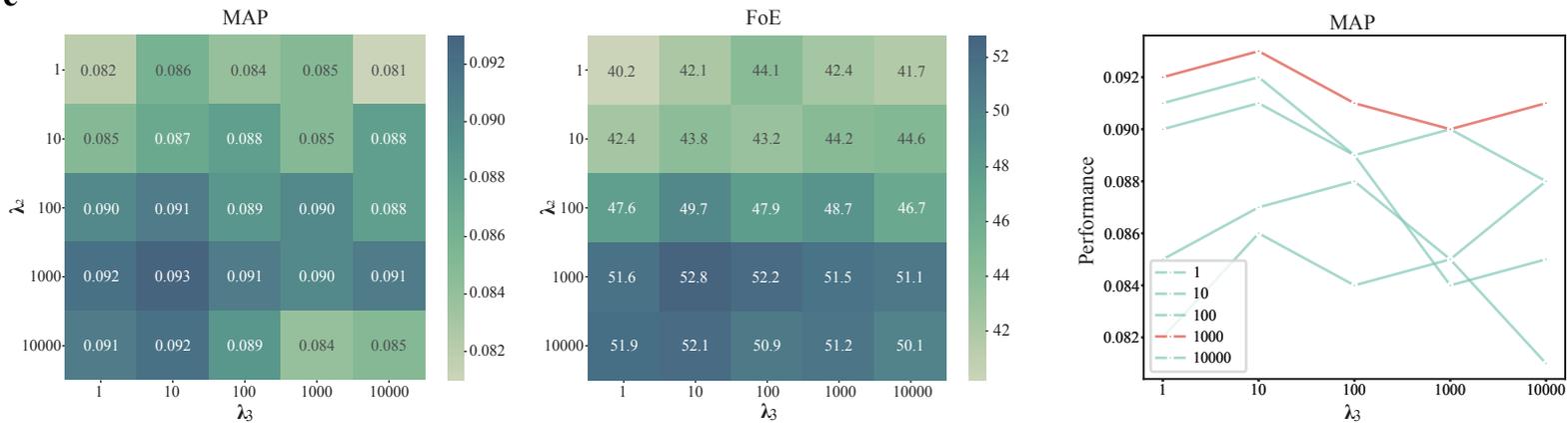

**Fig. 3**

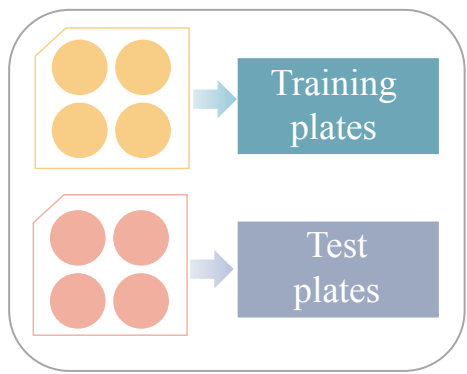
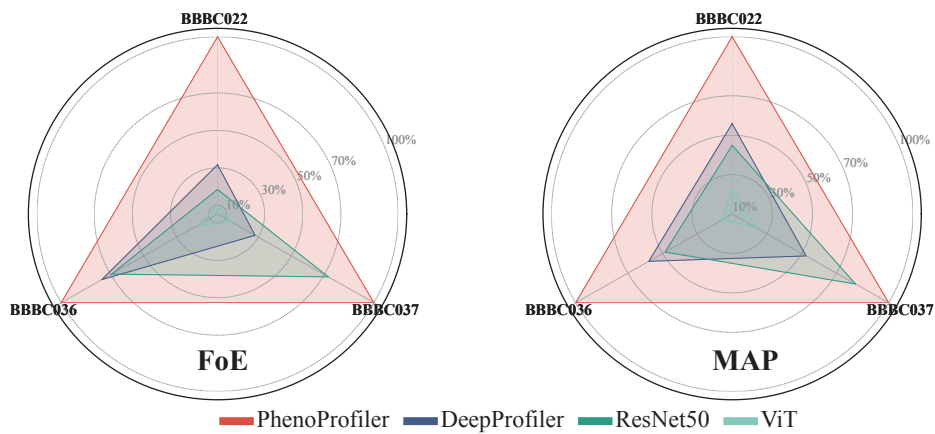

a  Leave-plates-out validation

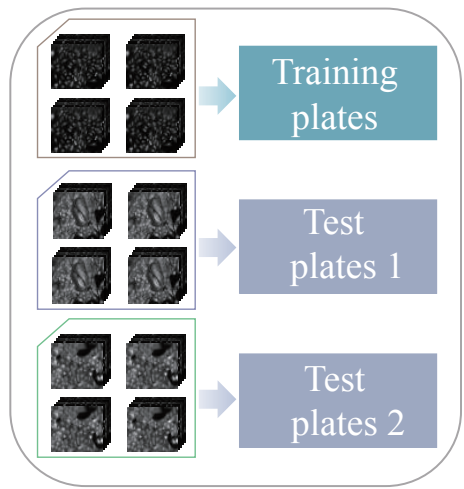
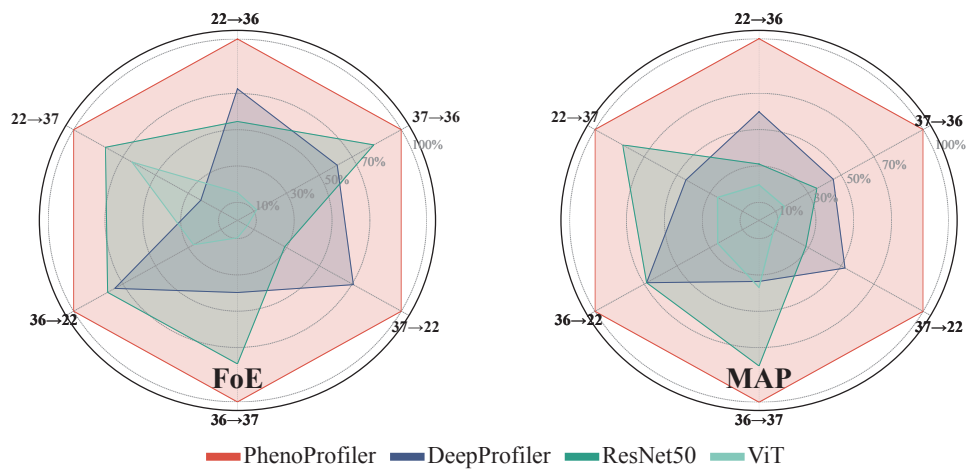

b  Leave-datasets-out validation

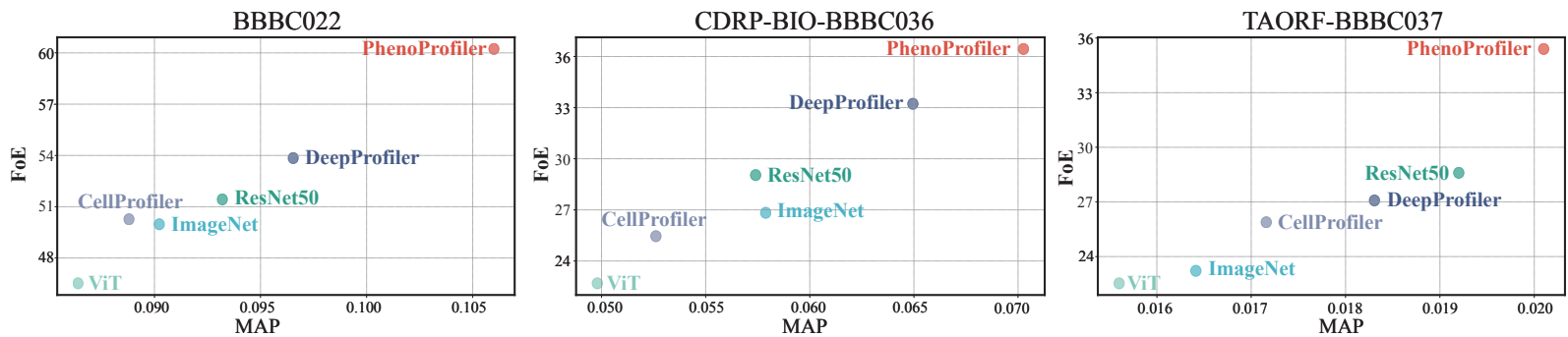

c  Non-end-to-end performance of phenotypic representations

**Fig. 4**

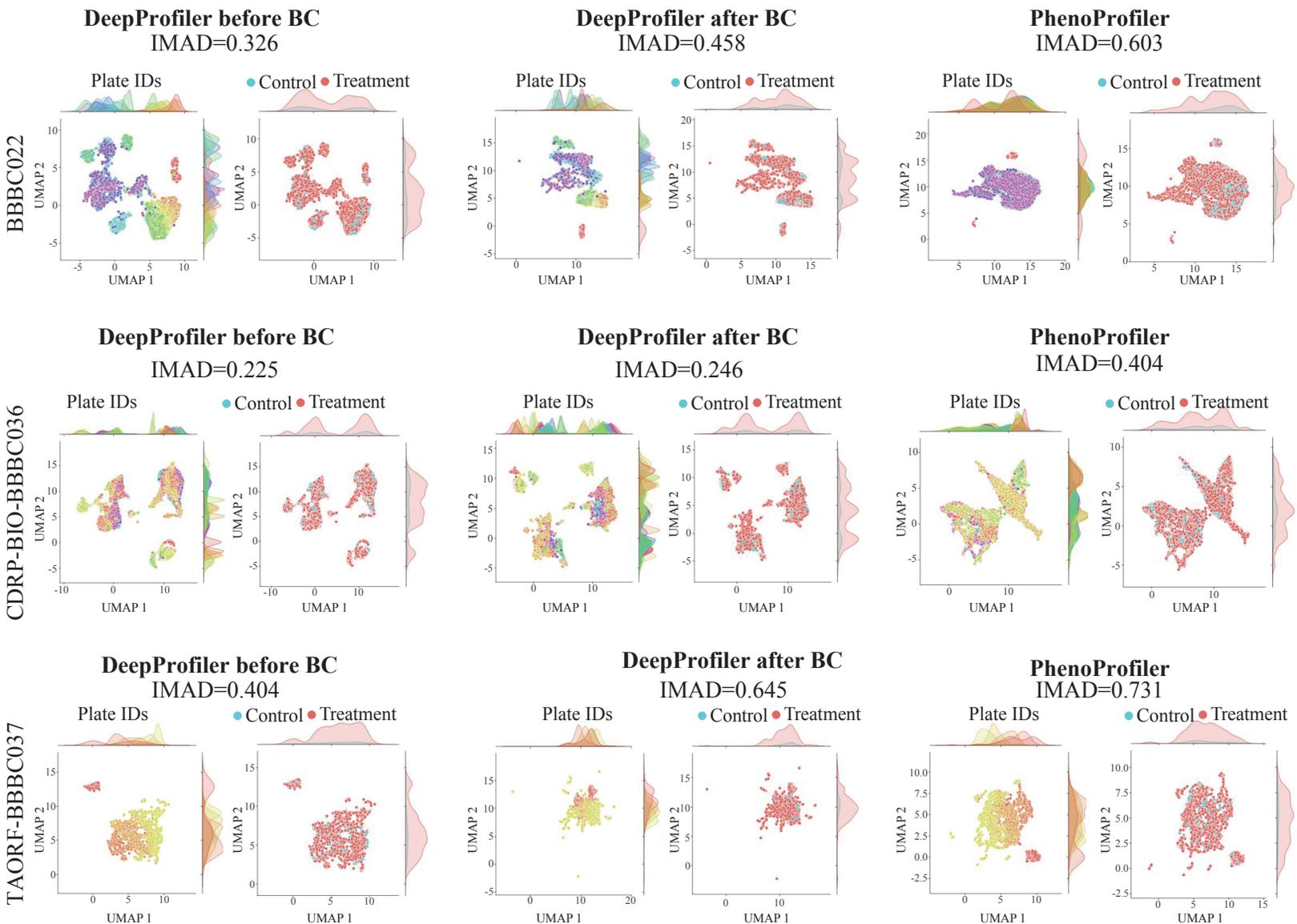

**Fig. 5**

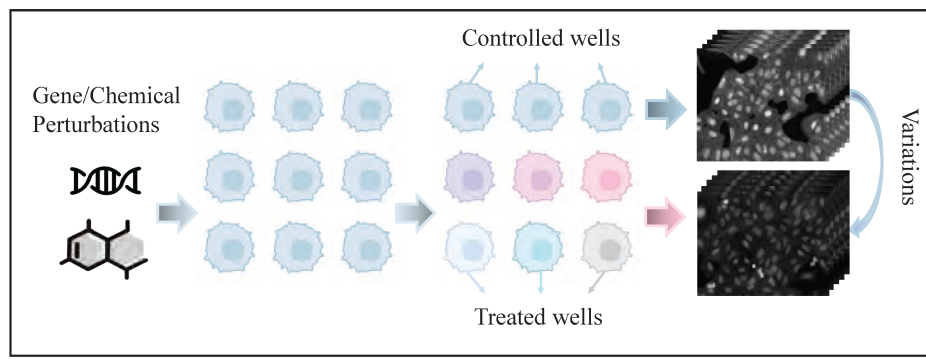
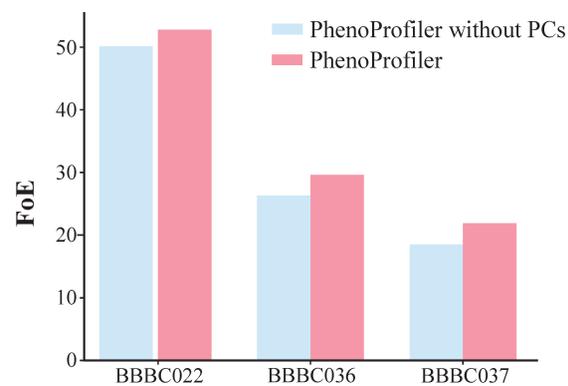
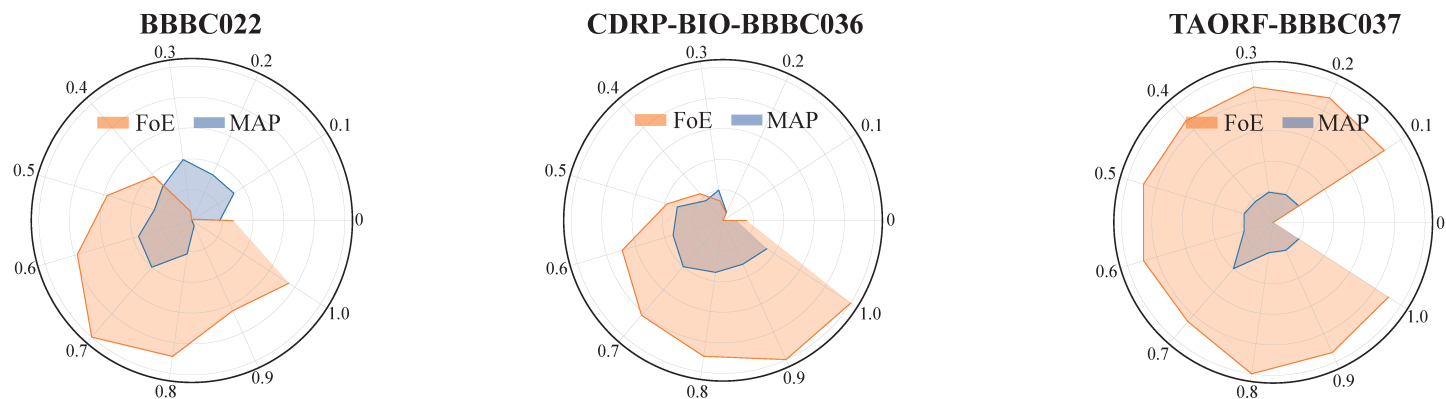
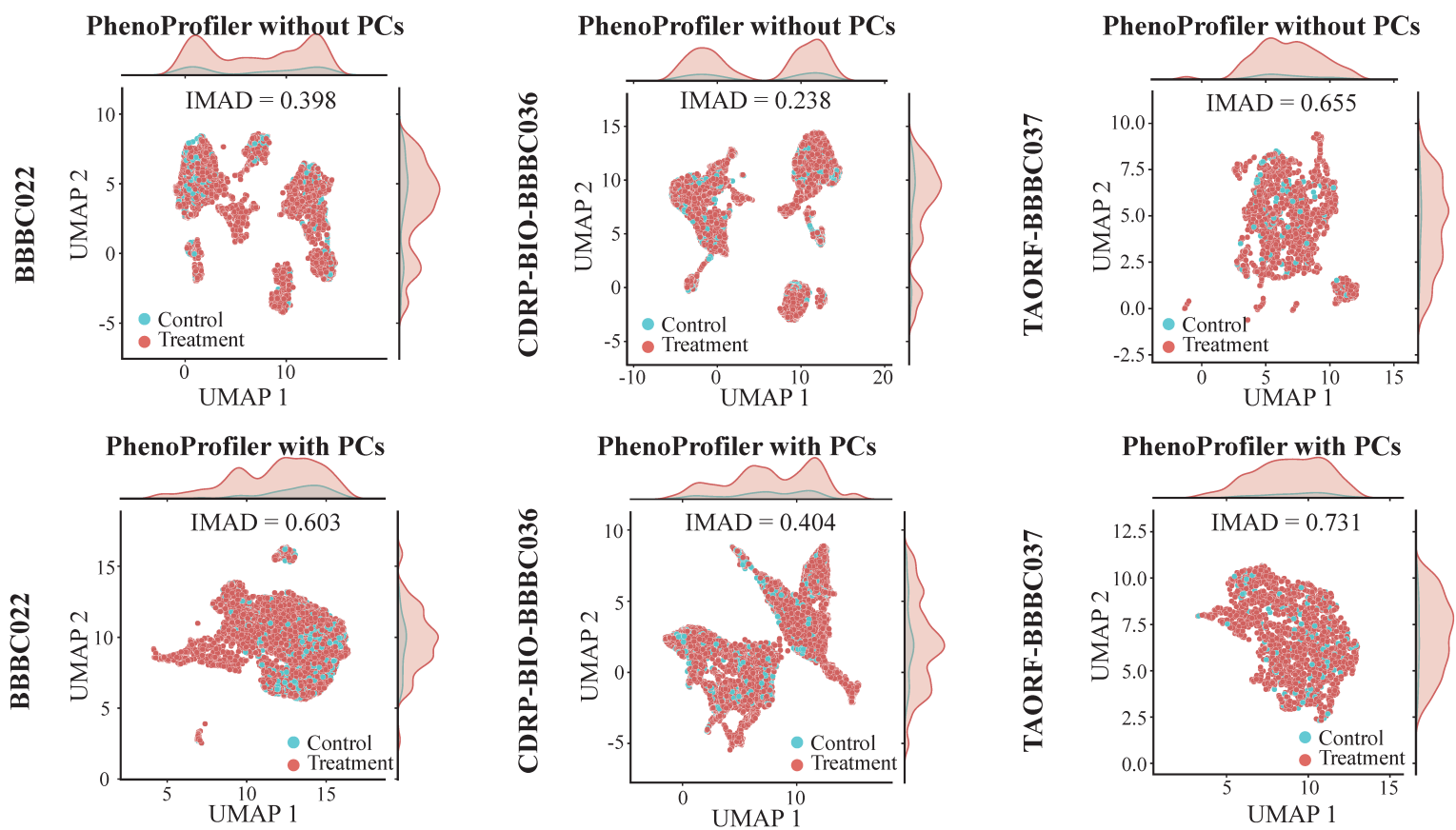

**Fig. 6**

**a**

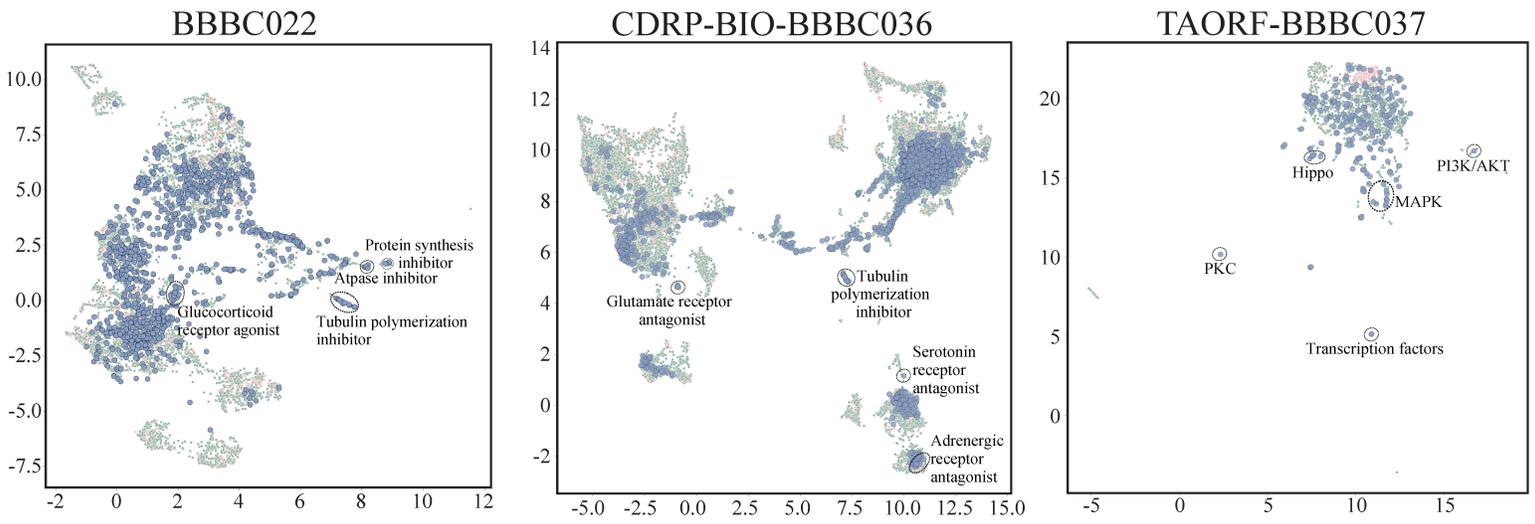

**b**

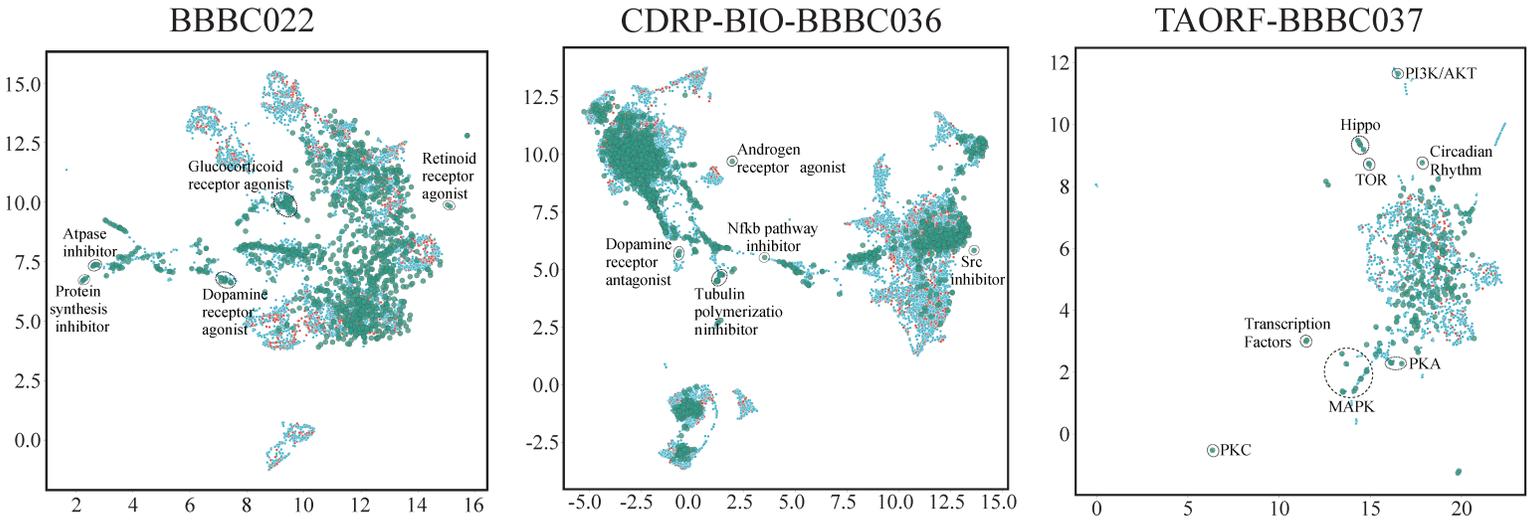